# Jal Anveshak: Prediction of fishing zones using fine-tuned LlaMa 2

Arnav Mejari[a], Maitreya Vaghulade[a],
Paarshva Chitaliya[a], Arya Telang[a] and Prof. Lynette D'mello[a]

*[a]Department of Computer Engineering, Dwarkadas J. Sanghvi College
of Engineering, JVPD Scheme, Mumbai, 400056, Maharashtra, India.*

**Abstract**

In recent years, the global and Indian government efforts in monitoring and collecting data related to the fisheries industry have witnessed significant advancements. Despite this wealth of data, there exists an untapped potential for leveraging artificial intelligence based technological systems to benefit Indian fishermen in coastal areas. To fill this void in the Indian technology ecosystem, the authors introduce Jal Anveshak. This is an application framework written in Dart and Flutter that uses a Llama 2 based Large Language Model fine-tuned on pre-processed and augmented government data related to fishing yield and availability. Its main purpose is to help Indian fishermen safely get the maximum yield of fish from coastal areas and to resolve their fishing related queries in multilingual and multimodal ways.

## 1. Introduction

Potential Fishing Zones (PFZs) is a unique service provided by the Indian National Centre for Ocean Information Services (INCOIS), that combines the efforts of oceanographers, remote sensing/GIS experts, and marine biologists. They are issued three times a week by INCOIS by incorporating the data on Sea Surface Temperature (SST) and Chlorophyll retrieved regularly from NOAA-AVHRR (USA), Eumetsat (ESA) Met-Op series satellites along with Oceansat-II (India) and MODIS Aqua (USA) satellites.

Shailesh Nayak et al. in a 2002 paper regarding the satellite based PFZ advisory service, highlight the socioeconomic benefits gained using the service. They estimated a saving of about Rs. 545 crores from the ten percent of boat owners that used the advisories in the same year as well as a predicted growth of Rs.1635 crores from about 25 percent of saving and boat owners respectively. These predictions indicate that the PFZ advisory serves to benefit the fishery industry as a whole, by saving the costs incurred by the fishermen for fuel as well as increasing the catch per unit effort. The data provided also gives information about the relative location of given ports albeit the ports keep changing as well as the data is difficult to understand for smaller fishermen. Another part missing from the PFZ data is the availability of location specific data as they only provide the name of ports, and a general latitude and longitude.

This research focuses on improving these advisories, by parsing the INCOIS data in a mobile app named "Jal Anveshak" and making it more interactive as well as providing location specific information to the fishmongers. In this app, a pre-trained and fine-tuned LLM is used to help navigate the fisherman to the nearest predicted fishing spot according to that day's data. The answer can be represented in either textual instructions or a GIS representation extracted from the model. By providing an easy to understand lower level abstraction of the entire data based on the location of the fisherman as well as their needs. This will help increase the usage of the data as from the current estimate of 25 percent as a raw data file fails to provide the layman with important data regarding fishing zones.

**Nomenclature**

| | |
|---|---|
| INCOIS | Indian National Centre for Ocean Information Services |
| PFZ | Potential Fishing Zones |
| SST | Sea Surface Temperature |

## 2. Literature Review

Kumar, T. & Masuluri, Nagaraja & Nayak, Shailesh. (2008) developed a model using sea surface temperature gradient, gradient persistence, and chlorophyll levels to characterize potential fishing zones by expected catch. Our app integrates a similar predictive capability, aiding fishermen in planning trips and managing expectations. This methodology enhances our fishing zone identification services, validated by the referenced paper.

Dwivedi, R.M. et al. (2005) demonstrate satellite data techniques for identifying potential fishing zones, with a high success rate in predicting productive areas and increased catch. Integrating satellite-based forecasts into our app provides valuable input for fishermen, improving benefit-cost ratios, as noted in the paper.

In their paper, Dwi Ely Kurniawan, Afdhol Dzikri, and Nanna Suryana Herman stress the importance of a tailored mobile app for fishermen to enhance catch marketing and sales. They recommend SMS gateway integration and consider infrastructure constraints for fishing zone identification. Incorporating user feedback and prototyping adjustments could enhance app effectiveness.

Akanmu Semiu Ayobami and Wan Rozaini Sheikh Osman in their paper offer insights into fishermen's essential app needs like weather updates and pricing information. They advocate participatory design involving fishermen, informing app development to meet genuine user needs.

Julia Calderwood in her paper reviews smartphone apps in commercial fisheries, stressing their role in data collection, including catch data for fishing zones. Co-design with fishermen is crucial for usability. Involving them ensures meeting needs, informing our app development.

Christian Skov, Kieran Hyder and Casper Gundelund in their paper highlight fishing apps as valuable tools for data collection, offering detailed spatiotemporal fishing behavior data.

T.M. Balakrishna Nair, K.Srinivas, M. Nagarajakumar and Dr. R. Harikumar underscore the advancements made by INCOIS in utilizing satellite observation systems to disseminate marine fishery advisories and potential fishing zone information. The datasets discussed in the paper support researchers in enhancing fishing conditions by providing relevant features for their methodologies.

In their paper , Majumder, Swarnali & Maity, Sourav & et.al  characterize Potential Fishing Zones (PFZs) along the Indian coast using machine learning. A hybrid decision tree model categorizes PFZs by fish catch quantity, offering potential for improved fishing zone recommendations.

H.U. Solanki, P.C. Mankodi, S.R Nayak and V.S. Somvanshi [9] study the relationship between fish diet, feeding habitat, and satellite-derived parameters to understand how satellite-based Potential Fishing Zones (PFZs) correlate with fishery assets. They calculate species-wise catch per unit effort (CPUE) and compare seasonal CPUE in PFZs with other areas within distinct ecosystems.

Trio Adiono , Febri Dawani , Erick Adinugraha & Aditia Rifai [10], propose a Long-Radio technology-based network as an alternative to LTE for connecting fishermen to shore at all ocean depths. Supporting up to 10 km from shore, it enables efficient transmission of SOS signals, weather updates, and live chat for fishermen.

In their paper, S. Jagannathan, A. Samraj and M. Rajavel [11], analyze INCOIS data from 2003 to 2011 for the Colachel coast, clustering based on factors like distance, oceanic depth, and wind direction. Fishing spots are categorized by season and wind direction from 165 spots, revealing potential spots between PFZ Points using latitude and longitude data.

To come up with an effective way to let large language models utilize the knowledge base fed to them, Patrick Lewis et al. [12] introduce retrieval-augmented generation, blending pre-trained parametric and non-parametric memory models to enhance knowledge-intensive NLP tasks. This approach surpasses previous LLM architectures by providing factual information beyond parametric baselines.

Shamane Siriwardhana, Rivindu Weerasekera and et al. [13] propose RAG-end2end for open-domain question answering, enabling domain-specific knowledge adaptation during training. Through joint training of generator and retriever, it covers datasets from various domains using the Huggingface Transformers library.

Xi Victoria Lin, Xilun Chen et al. [14] present Retrieval-Augmented Dual Instruction Tuning (RA-DIT), adding retrieval capabilities to LLMs through minimal fine-tuning. Optimizing for both contextual awareness and knowledge retrieval stages yields significant performance gains.

Humza Naveed, Asad Ullah Khan, Shi Qiu, Muhammad Saqib, Saeed Anware, Muhammad Usman et al. [15] comprehensively review LLM concepts, architectures, and challenges, addressing tokenization, attention, model configurations, computational costs, and overfitting.

Cheonsu Jeong [16] discusses LLM trends, pretraining, fine-tuning, and advancements like Parameter-efficient Fine-Tuning and adapter methods. The paper focuses on finance datasets, model selection, hyperparameter tuning, and evaluation.

Boqi Chen, Fandi Yi & Dániel Varró [17], propose a prompting approach for taxonomy constructions, outperforming fine-tuning in scenarios with limited information, reducing data quantity constraints and manual hierarchical relationship constructions.

D.S. Wang [18] presents an automated question answering system leveraging domain ontology and question templates, providing answers to user queries without deep natural language understanding. Matching user queries to predefined templates captures the user's intention effectively.

In this rapidly evolving landscape of fishing applications, Jal Anveshak stands out due to its distinct characteristics and capabilities. An evaluation of the parameters that characterize its functionality and user experience is necessary to determine where it stands in relation to other widely used fishing applications. In addition to highlighting our app's revolutionary features, this review offers a comprehensive comparison with other market leaders such as Fishbrain, FishAngler, ANGLR, and Smart Fishing Spots.

The criteria for evaluation include the prediction of fishing zones, data sources, real-time updates, offline support, user interaction, security, government collaboration, customization, user-friendliness, and cost. The primary features and value propositions that these apps provide are reflected in each of these criteria. For instance, Jal Anveshak's use of advanced AI and multiple satellite data sources for fishing zone prediction sets a new benchmark in the industry.

| Evaluation Criteria | Jal Anveshak | Fishbrain | FishAngler | ANGLR | Smart Fishing Spots |
|---|---|---|---|---|---|
| Prediction of Fishing Zones | ✓(Advanced AI, fine-tuned LLaMa 2) | ✓(AI-powered fishing spot prediction) | ✓(Real-time fishing forecasts) | ✓(Automated data capture) | ✓(AI & human intelligence) |
| Data sources | ✓(Multiple satellite sources: INCOIS, NOAA-AVHRR, Eumetsat, Oceansat-II, MODIS Aqua) | ✓(Crowdsourced data, Navionics charts) | ✓(Community data, weather forecasts) | ✓(USGS, Dark Sky) | ✓(Tides, weather, wind, barometric pressure) |
| Real-time updates | ✓ | ✓ | ✓ | ✓ | ✓ |
| Offline support | ✓ | ✗ | ✗ | ✗ | ✗ |
| User interaction | ✓(Text, voice, image) | ✗ | ✗ | ✗ | ✗ |
| Government Collabaration | ✓(Planned) | ✗ | ✗ | ✗ | ✗ |
| Customization | ✓(Multilingual, Indian region-specific) | ✗ | ✗ | ✗ | ✗ |
| User interface | ✓(User friendly, interactive maps and guided routes) | ✓(User friendly, Community-focused) | ✗ | ✗ | ✓(User friendly, interactive maps) |
| Cost | Free | Free basic version, Premium $5.99/month | Free | Free, with paid upgrades | Subscription-based |

## 3. Dataset Description

The paper mainly works with data scraped from the official government INCOIS website. Each day, there is an advisory about the probable location and their closest ports. As the oceanic conditions are dynamic, the potential fishing zone change and along with them so do the ports. This information is scraped everyday and stored along with relevant weather data such as humidity, sea surface temperature, etc. The weather data is collected using WeatherAPI, according to the latitude and longitude provided by the INCOIS advisory. The collection of data scraped for over a week acts as the initial dataset, to train the model; while the data scraped everyday in order to serve the relevant locations to the fishermen in conjunction with the validation of the same data done by the fishermen through the app is used to dynamically update the dataset according to weather conditions and the actual data.

The actual data collected by satellites is not readily available as area beyond certain latitude and longitude is classified as Exclusive Economic Zone (EEZ) and is highly regulated as well as restricted in terms of fishing.

## 4. Application Framework

### *4.1. Choice of framework*

A software library that offers a basic structure to facilitate the creation of applications for a particular environment is known as an application framework. An application framework serves as the framework upon which an application is constructed. Reducing the general problems encountered in application development is the goal of building application frameworks. Code that is compatible with many application modules is used to do this. Application frameworks are utilized in different fields, such as web-based applications, in addition to the development of graphical user interfaces (GUIs).

It is imperative to first decide on a framework to develop an application in. The wrong choice of frameworks can lead to multiple roadblocks in the software engineering process for the developer, thus slowing down the software development process and causing multiple delays in the same. For Jal Anveshak, the authors choose to develop the application using Dart and Flutter, with the reasons for doing so enumerated in the following two paragraphs.

 The object-oriented, client-optimized programming language Dart is used to create cross-platform mobile and web applications. It has user-friendly offerings that are useful for both server and user operations. Dart is mostly used to construct front-end user interfaces for mobile applications.

Developing cross-platform applications is increasingly being done with Flutter. Developers can create native-like applications more quickly and at a cheaper cost using Flutter since it runs on a single codebase and renders into native code on every platform. iOS, Android, Windows, MacOS, and other platforms are compatible with Flutter apps. The framework of the Flutter technology is its most noticeable component.

### *4.2. Framework structure*

The application framework Jal Anveshak first asks the user (presumably, a fisherman) to choose a language from a dropdown in which they want the application to be in (See Fig. 1). All the internal text of the website from this point thereon is in the chosen language, which can be later changed internally in the settings if required (See Fig. 2).

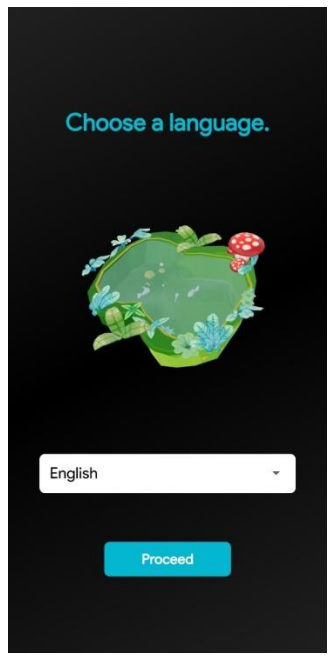
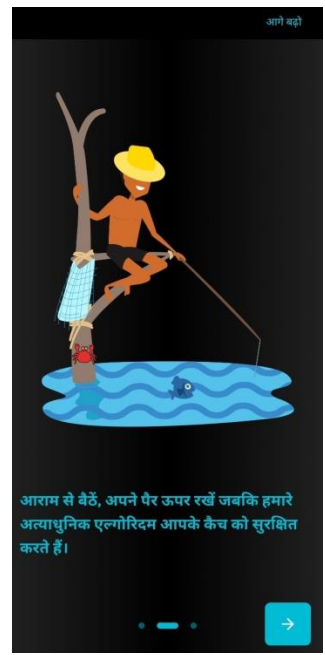

Fig. 1. Language Selection Page        Fig. 2. Multilingual Support

After the language is chosen, three skippable landing pages are shown that underline how the application aims to ease the lives of fishermen. Then, we have the login page. If the user does not have an account, we ask them to sign up with first name, last name, email, and password as inputs. Once the user signs up, we route them back to the login page, where they can login with their email and password. To ensure the utmost safety of data and security of the fisherman's details in the application, we further request fingerprint authentication as well if the entered combination of email and password are found to be valid. If the user has forgotten their password, they can reset it via email. If the fingerprint is authenticated, the application logic leads the user inside the application.

Here, the user is afforded three modes of doubt resolution: one via simple text input, another via voice input and finally one via image upload/camera. All of these connect with the machine learning models (explained in the above sections) in the backend and give users appropriate answers for their queries. The app provides an option to enable dark mode to maintain a certain standard of ease of use in terms of user interaction and user experience.

The authors also have implemented a maps feature, which fulfills the main criteria of the project: predicting potential fishing zones. It asks permission to use the location of the mobile phone or smartphone being used and then puts down markers on the most appropriate fishing zones for the user in that location on the rendered map, thus helping the fisherman determine the spots to which they should go to for getting maximum yield of fishes with safety and weather also being considered.

The app also possesses an offline SOS calling feature. On voice input of a "wake word" that activates this calling feature, a voice call is made to the Coast Guard helpline so as to alert them to any possible mishaps or dangers a fisherman could be on their fishing vessel or boat. This feature adds a layer of security to the fisherman's daily operations without utilizing any APIs or the internet.

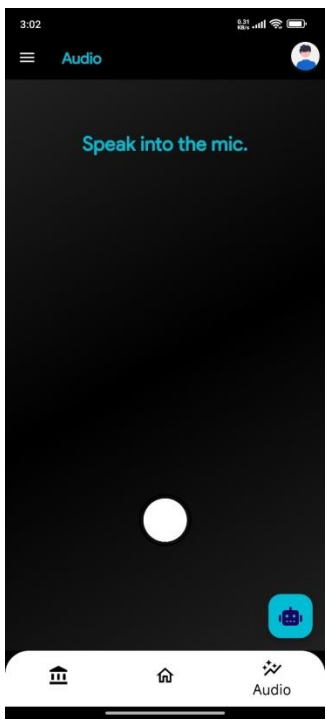

Fig. 3. Multimodal Capability

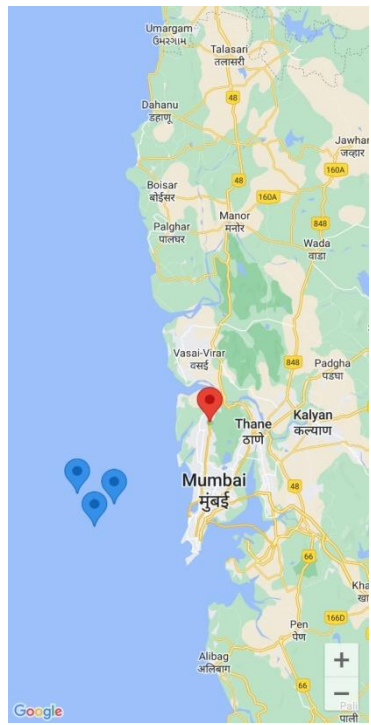

Fig. 4. Fishing Zone Markers

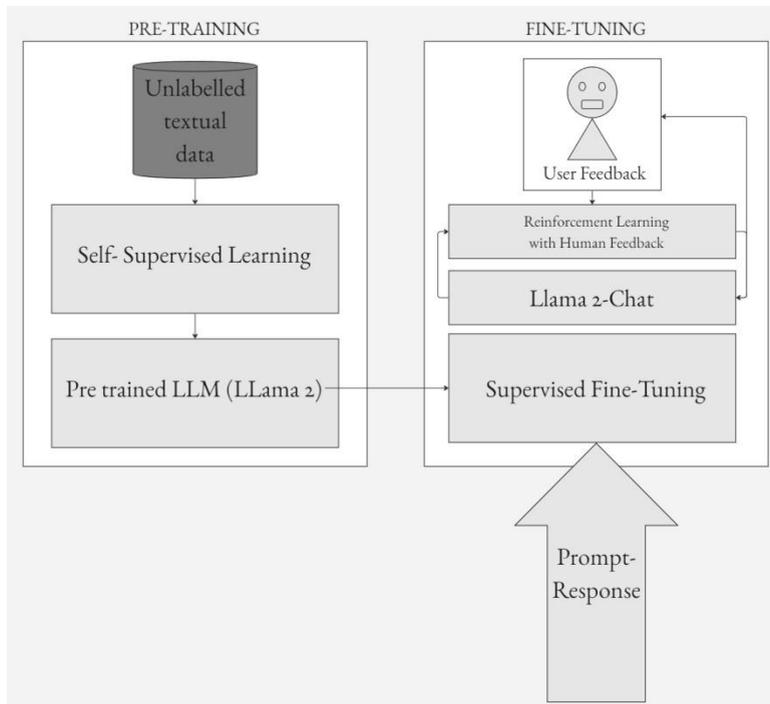

Fig. 5. Training and Pre-training of LLM

*4.3. NLP Methodology*

Large language models (LLMs) hold great promise as intelligent AI helpers that can handle difficult thinking tasks in a variety of fields. Because these models allow for conversational interfaces that are easy to use, the general public has embraced them quickly and extensively and that is the exact reason why we decided to incorporate a large language model in our app.

To improve the Llama 2 model on the fishing domain, we created a dataset with 10,000 entries. The monthly data was gathered on a daily basis, and each day's data was matched with a pre-established set of questions and their corresponding responses. The dataset was painstakingly assembled. The Llama tokenizer was used to tokenize the data, guaranteeing that it was correctly encoded for additional processing, in order to get the dataset ready for model training. The dataset was tokenized and then entered into the module known as SFTTrainer, or Supervised Fine-tuning Trainer. This module made the process of fine-tuning easier, enabling the model to efficiently learn from the encoded dataset. Under supervised conditions, the model's performance was fine-tuned to maximize its effectiveness on the given task.

The purpose of this fine-tuned chatbot was to make the data more accessible to the fishermen, improve doubt resolution capabilities and enhance user friendliness. For example, the model can answer the question "Where should I go to catch Rohu fish today?" with accurate detail through text and markers in interactive maps nearest to the fisherman's location (See Fig. 4).

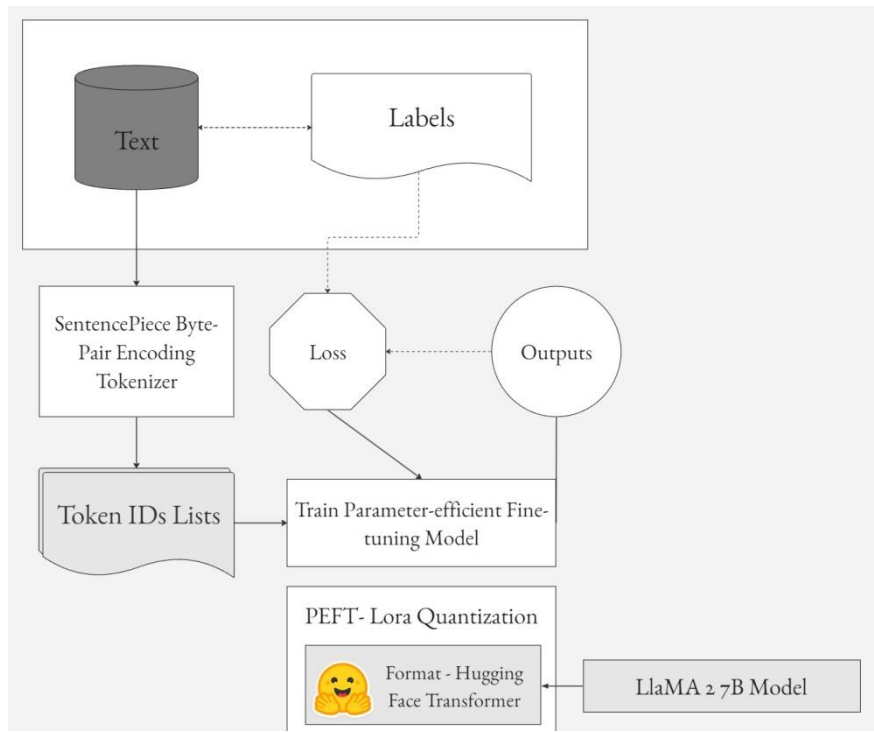

Fig. 6. Finetuning LlaMA 2 7B Model

## 5. Future Work

The proposed work could have future enhancements by implementing analysis of satellite images and breed specific fish data in order to better suit the needs of fishermen. Additionally, making the app accessible without network connectivity would provide better offline support to offshore fishermen. Further, involving the necessary government agencies, will help in promoting the product and in its recognition. With the help of the government, the app will be made official, hence reaching to the target audience, i.e. the fishermen.